\documentstyle[11pt]{article}

\input{epsf}

\newcommand{\fig}[3]
{\begin{figure}[htb]
\centerline{
\epsfxsize #2
\epsffile{#1}}
\caption{#3}
\end{figure}
}

\begin{document}

\title{Memory-Based Lexical Acquisition and Processing}

\author{Walter Daelemans \thanks{Preprint of a paper to be published
in Steffens (ed.) {\em Machine Translation and the Lexicon.} Springer
Lecture Notes in Artificial Intelligence.  I would like to thank my
colleagues in the {\it Atila} project (Steven Gillis, Gert Durieux,
and Antal van den Bosch) for their contributions to the approach
described in this paper. The {\it Atila} (Antwerp-Tilburg Inductive
Language Acquisition) project is a research corporation between the
University of Antwerp and Tilburg University focusing on the
application of Machine Learning techniques in linguistic engineering
and in developmental psycholinguistics. Thanks also to the
participants of the Heidelberg workshop on Machine Translation and the
Lexicon for useful comments and suggestions.}}

\date{Institute for Language Technology and AI, Tilburg
University\\
P.O.Box 90153, 5000 LE Tilburg, The Netherlands\\
Walter.Daelemans@kub.nl}

\maketitle

\begin{abstract}
Current approaches to computational lexicology in language technology
are knowledge-based (competence-oriented) and try to abstract away
from specific formalisms, domains, and applications. This results in
severe complexity, acquisition and reusability bottlenecks.  As an
alternative, we propose a particular performance-oriented approach to
Natural Language Processing based on automatic memory-based learning
of linguistic (lexical) tasks.  The consequences of the approach for
computational lexicology are discussed, and the application of the
approach on a number of lexical acquisition and disambiguation tasks
in phonology, morphology and syntax is described.
\end{abstract}

\section{Introduction}

In computational lexicology, three basic questions guide current
research: (1) which knowledge should be in the lexicon, (2) how should
this knowledge be represented (e.g., to cope with the problems of
lexical gaps), and (3) how can this knowledge be acquired. Current
lexical research in language technology is eminently {\it
knowledge-based} in this respect.  It is also generally acknowledged
that there exists a natural order of dependencies between these three
research questions: acquisition techniques depend on the type of
knowledge representation used and the type of knowledge that should be
acquired, and the type of knowledge representation used depends on
what should be represented.

Also uncontroversial, but apparently no priority issue for many
researchers, is the fact that the question which knowledge should be
represented (which morphological, syntactic, and semantic {\it senses}
of lexical items should be distinguished, \cite{pust})
depends completely on the natural language processing {\it task} that
is to be solved.  Different tasks require different lexical
information. Also, different theoretical formalisms, domains, and
languages require different types of lexical information and therefore
possibly also different types of lexical knowledge representation and
different acquisition methods.  It makes sense to work on ``a lexicon
for HPSG parsing of Dutch texts about airplane parts'' or on
``lexicons for translating computer manuals from English to Italian'',
but does it make equal sense to work on ``the lexicon''?  Because it
is uncontroversial that lexicon contents is a function of task,
domain, language, and theoretical formalism, the {\it reusability
problem} has been defined as an additional research topic in
computational lexicology, an area that should solve the problem of how
to translate lexical knowledge from one theory, domain, or application
to the other. Unfortunately, successful solutions are limited and few.

In this paper, we propose an alternative approach in which a
performance-oriented (behaviour-based) perspective is taken instead of
a competence-oriented (knowledge-based) one.  We try to automatically
{\it learn} the language processing task on the basis of examples. The
effect of this is that the priorities between the three goals
discussed earlier are changed: the representation of the acquired
knowledge depends on the acquisition technique used, and the knowledge
acquired depends on what the learning algorithm has induced as being
relevant in solving the task.  This shift in focus introduces a new
type of reusability: reusability of {\it acquisition method} rather
than reusability of acquired knowledge. It also has as a consequence
that it is no longer a priori evident that there should be different
components for lexical and non-lexical knowledge in the internal
representation of an NLP system solving a task, except when the task
learned is specifically lexical.

The structure of the paper will be as follows. In Section 2 we will
explain the differences between the knowledge-based and the
behaviour-based approach to Natural Language Processing (NLP). Section
3 introduces {\it lazy learning}, the symbolic machine learning
paradigm which we have used in experiments in lexical acquisition. In
Section 4, we show how virtually all linguistic tasks can be redefined
as a classification task, which can in principle be solved by lazy
learning algorithms.  Section 5 gives an overview of research results
in applying lazy learning to the acquisition of lexical knowledge, and
Section 6 concludes with a discussion of advantages and limitations of
the approach.

\section{Knowledge-Based versus Behaviour-Based}

One of the central intuitions in current knowledge-based NLP research
is that in solving a linguistic task (like text-to-speech conversion,
parsing, or translation), the more linguistic knowledge is explicitly
modeled in terms of rules and knowledge bases, the better the
performance.

As far as lexical knowledge is concerned, this knowledge is
represented in a lexical knowledge base, introduced either by hand or
semi-automatically using machine-readable dictionaries.  The problem
of reusability is dealt with by imposing standards on the
representation of the knowledge, or by applying filters or translators
to the lexical knowledge. Not only is there a huge and costly {\it
linguistic engineering} effort involved in the building of a
knowledge-based lexicon in the first place, the effort is duplicated
for every translation module between two different formats of the
lexical knowledge. In practice, most NLP projects therefore start
lexicon construction from scratch, and end up with unrealistically few
lexical items.

In this paper, we will claim that regardless of the state of
theory-formation about some linguistic task, simple data-driven
learning techniques, containing very little a priori linguistic
knowledge, can lead to performance systems solving the task with an
accuracy higher than state-of-the art knowledge-based systems. We will
defend the view that all linguistic tasks can be formulated as a {\it
classification} task, and that simple memory-based learning techniques
based on a {\it consistency heuristic} can learn these classifications
tasks.

\begin{quote}
{\bf Consistency Heuristic}. ``Whenever you want to guess a property
of something, given nothing else to go on but a set of reference
cases, find the most similar case, as measured by known properties,
for which the property is known. Guess that the unknown property is
the same as that known property.'' (\cite{winst})
\end{quote}

In this approach, reusability resides in the {\it acquisition method}.
The same, simple, machine learning method may be used to induce
linguistic mappings whenever a suitable number of examples (a corpus)
is available, and can be reused for any number of training sets
representing different domains, sublanguages, languages, theoretical
formalisms, and applications.  In this approach, emphasis shifts from
knowledge representation (competence) to induction of systems exposing
useful behaviour (performance), and from knowledge engineering to the
simpler process of data collection. Fig. 1 illustrates the
difference between the two approaches.


\fig{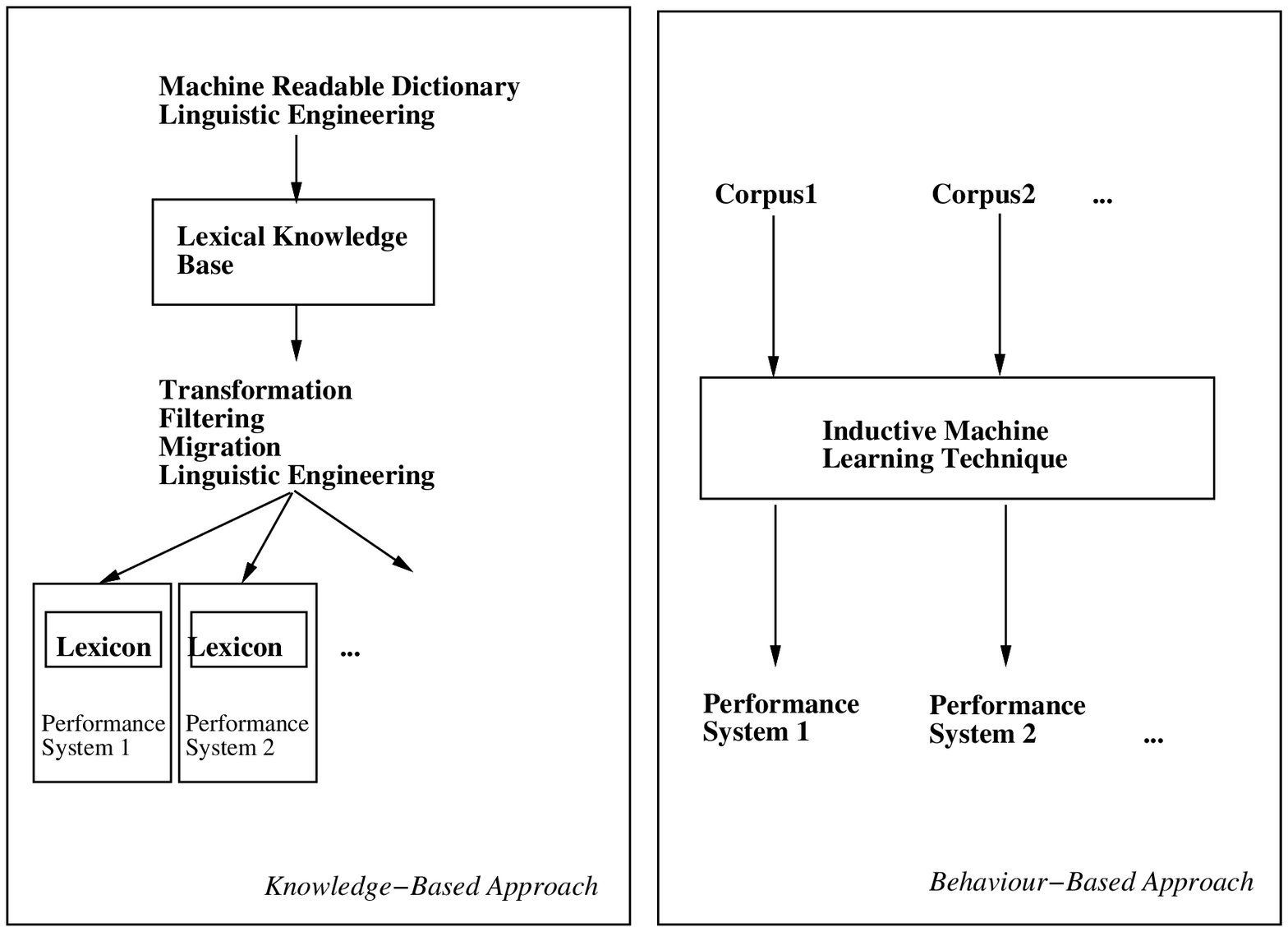}{11cm}{Knowledge-Based versus Behaviour-Based approaches
to lexical acquisition}

\section{Supervised Machine Learning of Linguistic Tasks}

In supervised Machine Learning, a learner is presented with a number
of examples describing a mapping to be learned, and the learner should
extract the necessary regularities from the examples and apply them to
new, previously unseen input.  It is useful in Machine Learning to
make a distinction between a {\it learning component} and a {\it
performance component}.  The performance component produces an output
(e.g., a syntactic category) when presented with an input (e.g., a
word and its context) using some kind of representation (decision
trees, classification hierarchies, rules, exemplars, \ldots). The
learning component implements a learning method. It is presented with
a number of examples of the required input-output mapping, and as a
result modifies the representation used by the performance system to
achieve this mapping for new, previously unseen inputs. There are
several ways in which {\em domain bias} (a priori knowledge about the
task to be learned) can be used to optimize learning. In the
experiments to be described we will not make use of this possibility.

There are several ways we can measure the success of a learning
method. The most straightforward way is to measure {\it accuracy}. We
randomly split a representative set of examples into a training set
and a test set\footnote{To have reliable results, this process is
repeated 10 times with different partitions of 90\% training and 10\%
test items, and the average success rate of these ten experiments is
computed (\cite{wei:kuli}).}, train the system on the training set,
and compute the success rate (accuracy) of the system on the test set,
i.e., the number of times the output of the system was equal to the
desired output. Other evaluation criteria include learning and
performance speed, memory requirements, clarity of learned
representations, etc.

\subsection{Lazy Learning}

Recently, there has been an increased interest in Machine Learning for
{\it lazy learning} methods.  In this type of similarity-based
learning, classifiers keep in memory (a selection of) examples without
creating abstractions in the form of rules or decision trees (hence
{\it lazy} learning).  Generalization to a new input pattern is
achieved by retrieving the most similar memory item according to some
distance metric, and extrapolating the category of this item to the
new input pattern (applying the consistency heuristic).  Instances of
this form of {\it nearest neighbour method} include instance-based
learning (\cite{aha:al}), exemplar-based learning (\cite{sal},
\cite{cos:sal}), memory-based reasoning (\cite{stan:wal}), and
case-based reasoning (\cite{kol}). Advantages of the approach include
an often surprisingly high classification accuracy, the capacity to
learn polymorphous concepts, high speed of learning, and perspicuity
of algorithm and classification (see e.g., \cite{cos:sal}).  Learning
speed is extremely fast (it consists basically of storing patterns),
and performance speed, while relatively slow on serial machines, can
be considerably reduced by using k-d trees on serial machines
(\cite{fried}), massively parallel machines (\cite{stan:wal}), or
Wafer-Scale Integration (\cite{kit}).  In Natural Language Processing,
lazy learning techniques are currently also being applied by various
Japanese groups to parsing and machine translation under the names
{\it exemplar-based translation} or {\it memory-based translation and
parsing} (\cite{kit}).

Lazy learning has diverse intellectual dependencies: in AI techniques
like memory-based reasoning and case-based reasoning, it is stressed
that ``intelligent performance is the result of the use of memories of
earlier experiences rather than the application of explicit but
inaccessible rules'' (\cite{stan:wal}). Outside the linguistic
mainstream, people like Skousen, Derwing, and Bybee stress that ``the
analogical approach (as opposed to the rule-based approach) should
receive more attention in the light of psycholinguistic results and
new formalizations of the notion of analogy'' (\cite{skou};
\cite{der:sko}), In cognitive psychology (e.g.,
\cite{smi:med}), exemplar-based categorization has a long history as
an alternative for probabilistic and classical rule-based
classification, and finally, in statistical pattern recognition, there
is a long tradition of research on {\it nearest neighbour}
classification methods which has been a source of inspiration for the
development of lazy learning algorithms.

\subsection{Variants of Lazy Learning}

Examples are represented as a vector of feature values with an
associated category label. Features define a pattern space, in which
similar examples occupy regions that are associated with the same
category (note that with symbolic, unordered feature values, this
geometric interpretation doesn't make sense).

During {\it training}, a set of examples (the training set) is
presented in an incremental fashion to the classifier, and added to
memory.  During {\it testing}, a set of previously unseen
feature-value patterns (the test set) is presented to the system. For
each test pattern, its distance to all examples in memory is computed,
and the category of the least distant instance is used as the
predicted category for the test pattern.

In lazy learning, performance crucially depends on the distance metric
used.  The most straightforward distance metric would be the one in
equation (1), where $X$ and $Y$ are the patterns to be compared, and
$\delta (x_{i},y_{i})$ is the distance between the values of the
$i$-th feature in a pattern with $n$ features.

\begin{equation}
\Delta(X,Y) = \sum_{i=1}^{n} \delta(x_{i},y_{i})
\end{equation}

Distance between two values is measured using (2) for numeric features
(using scaling to make the effect of numeric features with different
lower and upper bounds comparable), and (3), an overlap metric, for
symbolic features.

\begin{equation}
\delta(x_{i}, y_{i}) = \frac{|x_{i} - y_{i}|}{max_{i} - min_{i}}
\end{equation}

\begin{equation}
\delta(x_{i}, y_{i}) = 0\mbox{ if }x_{i} = y_{i},\mbox{ else } 1
\end{equation}

\subsection{Feature weighting}

In the distance metric described above, all features describing an
example are interpreted as being equally important in solving the
classification problem, but this is not necessarily the case.
Elsewhere (\cite{dae:bos:a}; \cite{dae:al:a}) we introduced the
concept of information gain (also used in decision tree learning,
\cite{quin}) into lazy learning to weigh the importance of
different features in a domain-independent way. Many other methods to
weigh the relative importance of features have been designed, both in
statistical pattern recognition and in machine learning (e.g.,
\cite{aha};
\cite{kir:ren}; etc.), but the one we used is extremely simple and
produced excellent results.

The main idea of {\it information gain weighting} is to interpret the
training set as an information source capable of generating a number
of messages (the different category labels) with a certain
probability. The information entropy of such an information source can
be compared in turn for each feature to the average information
entropy of the information source when the value of that feature is
known. Those features that reduce entropy most are most informative.

Database information entropy is equal to the number of bits of
information needed to know the category given a pattern. It is
computed by (4), where $p_{i}$ (the probability of category $i$) is
estimated by its relative frequency in the training set.

\begin{equation}
H(D) = - \sum_{p_{i}} p_{i} log_{2} p_{i}
\end{equation}

For each feature, it is now computed what the information gain is of
knowing its value. To do this, we compute the average information
entropy for this feature and subtract it from the information entropy
of the database. To compute the average information entropy for a
feature (5), we take the average information entropy of the database
restricted to each possible value for the feature. The expression
$D_{[f=v]}$ refers to those patterns in the database that have value
$v$ for feature $f$. $V$ is the set of possible values for feature
$f$. Finally, $|D|$ is the number of patterns in a (sub)database.

\begin{equation}
H(D_{[f]}) = \sum_{v_{i} \in V} H(D_{[f=v_{i}]}) \frac{|D_{[f=v_{i}]}|}{|D|}
\end{equation}

Information gain is then obtained by (6), and scaled to be used as a
weight for the feature during distance computation.

\begin{equation}
G(f) = H(D) - H(D_{[f]})
\end{equation}

Finally, the distance metric in (1) is modified to take into account
the information gain weight associated with each feature.

\begin{equation}
\Delta(X,Y) = \sum_{i=1}^{n} G(f_{i}) \delta(x_{i},y_{i})
\end{equation}

Even in itself, information gain may be a useful measure to discover
which features are important to solve a linguistic task. Fig. 2 shows
the information gain pattern for the prediction of the diminutive
suffix of nouns in Dutch. In this task, features are an encoding of
the two last syllables of the noun the diminutive suffix of which has
to be predicted (there are five forms of this suffix in Dutch). Each
part (onset, nucleus, coda) of each of the two syllables (if present)
is a separate feature. For each syllable, the presence or absence of
stress is coded as well. The feature information gain pattern clearly
shows that most relevant information for predicting the suffix is in
the rime (nucleus and coda) of the last syllable, and that stress is
not very informative for this task (which conforms to recent
linguistic theory about diminutive formation in Dutch).


\fig{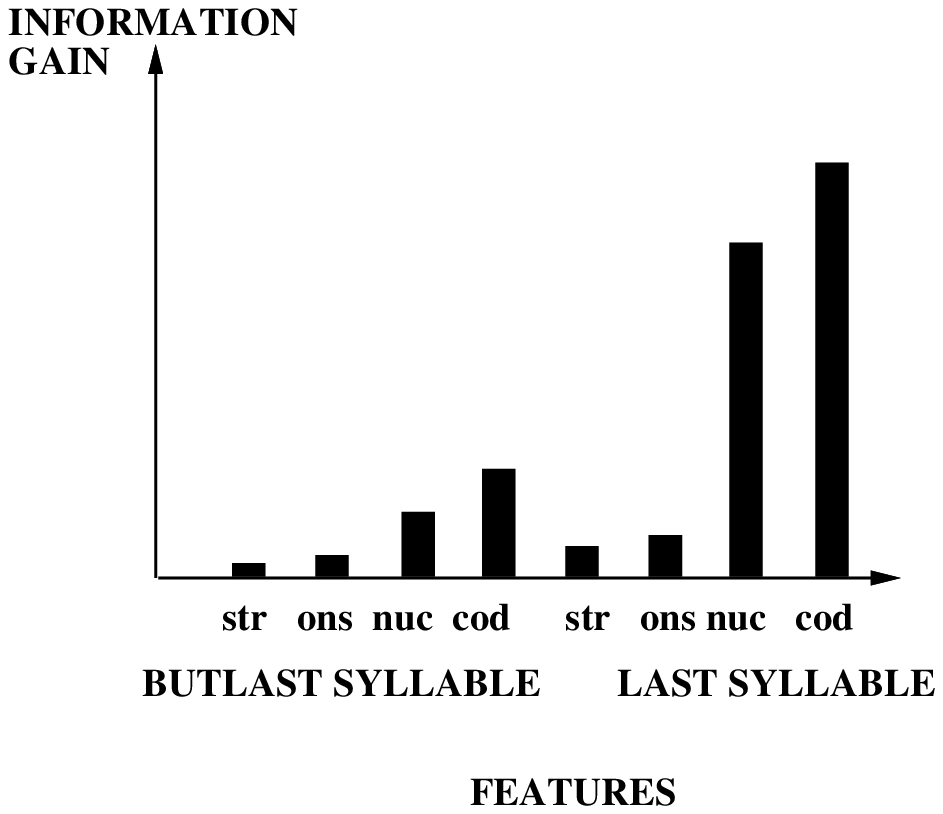}{8 cm}{An example of an information gain pattern. The
height of the bars expresses, for each feature describing an input
word, the amount of information gain it contributes to predicting the
suffix. Features are stress (str), onset (ons), nucleus (nuc), and
coda (cod) of the last two syllables of the noun.}

\subsection{Additional Extensions}

Apart from the feature weighting solution, several other optimizations
of the algorithm are possible. These concern, e.g., the use of
symbolic features: when using the previous metric, all values of a
feature are interpreted as equally distant to each other. This may
lead to unsufficient discriminatory power between patterns. It also
makes impossible the well-understood ``Euclidean distance in pattern
space'' interpretation of the distance metric. Stanfill and Waltz
(\cite{stan:wal}) proposed a {\it value difference metric} which takes
into account the overall similarity of classification of all examples
for each value of each feature.  Recently, Cost and Salzberg
(\cite{cos:sal}) modified this metric by making it symmetric.

In addition, the exemplars themselves can be weighted, based on
typicality (how typical is a memory item for its category) or
performance (how well is an exemplar doing in predicting the category
of test patterns), storage can be minimized by keeping only a
selection of examples, etc.

\section{Lazy Learning of Linguistic Tasks}

Linguistic tasks (including lexical tasks) are context-sensitive
mappings from one representation to another (e.g., from text to
speech, from spelling to parse tree, from parse tree to logical form,
from source language to target language etc.). These mappings tend to
be many-to-many and complex because they can often only be described
by conflicting regularities, sub-regularities, and exceptions.

In current NLP, these different levels of generalization have been the
prime motivation for research into inheritance mechanisms and default
reasoning (\cite{dae:gaz}; \cite{bris}), especially in research on the
structure and organisation of the lexicon.

To illustrate the difference between the traditional knowledge-based
approach with the lazy learning approach, consider Fig. 3. Suppose a
problem can be described by referring to only two features (a typical
problem would need tens or hundreds of features). In a knowledge-based
approach, the computational linguist looks for dimensions (features)
to describe the solution space, and formulates rules which in their
condition part define areas in this space and in their action part the
category or solution associated with this area.  Areas may overlap,
which makes necessary some form of rule ordering or ``elsewhere
condition'' principle.


\fig{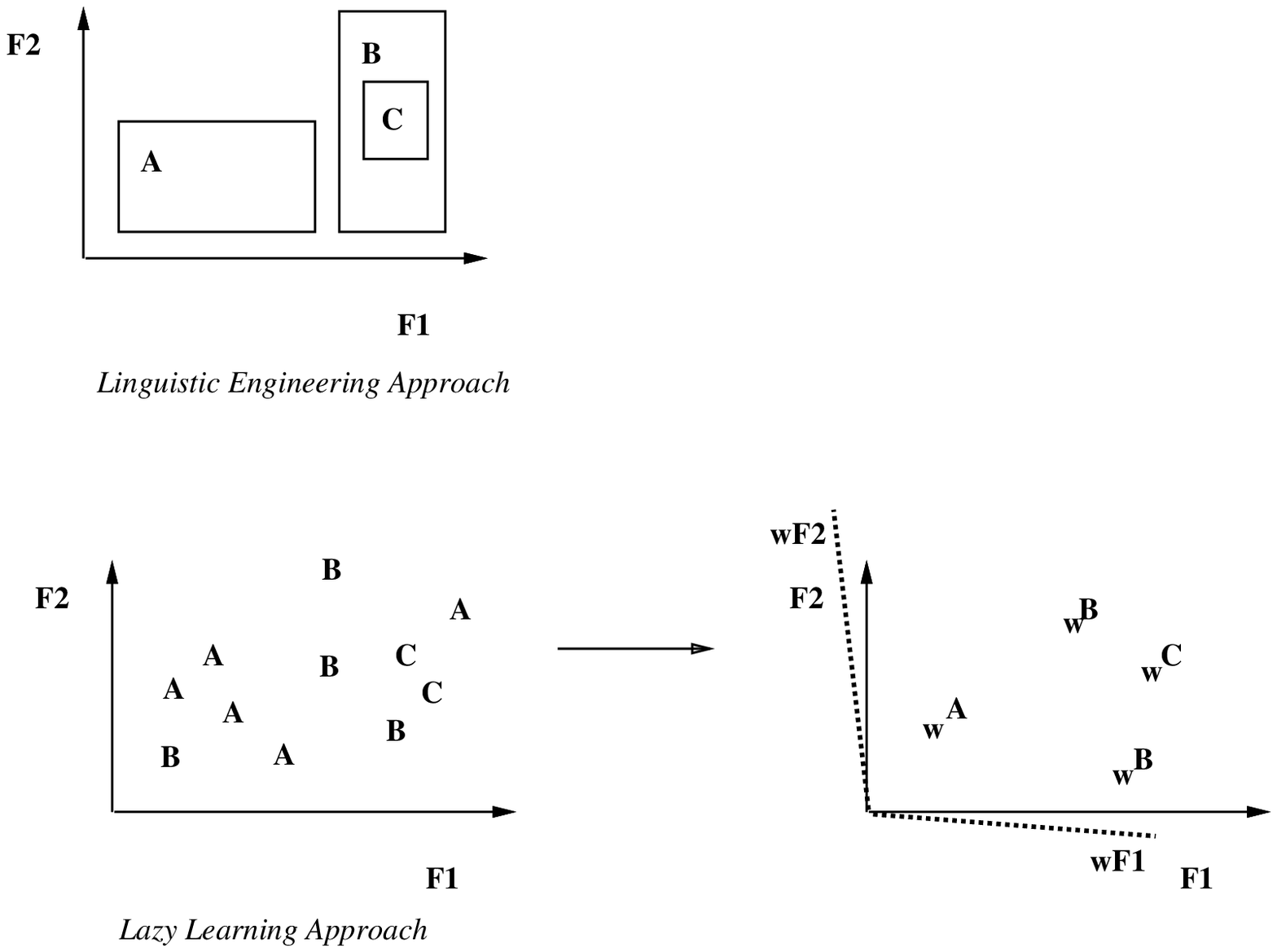}{11cm}{A graphical view of the difference between
linguistic engineering (top, knowledge-based) and lazy learning
(bottom, behaviour-based)}

For example, the two dimensions might be case and number of adjectives
in some language, and the three categories might be different suffixes
associated with different combinations of values for the case and
number features.

In a lazy learning approach, on the other hand, knowledge acquisition
is automatic. We start from a number of examples, which can be
represented as points in feature space. This initial set of examples
may contain noise, misclassifications, etc. Information-theoretic
metrics like information gain basically modify this feature space
automatically by assigning more or less weight to particular features
(dimensions). In constructive induction, completely new feature
dimensions may be introduced for separating the different category
areas better in feature space. Exemplar weighting and memory
compression schemes modify feature space further by removing points
(exemplars) and by increasing or decreasing the ``attraction area'' of
exemplars, i.e., the size of the neighbourhood of an exemplar in which
this exemplar is counted as the nearest neighbour. We are finally left
with a reorganized feature space that optimally separates the
different categories, and provides good generalization to unseen
inputs. In this process, no linguistic engineering and no handcrafting
were involved.

\subsection{Linguistic Tasks as Classification}

Lazy Learning is fundamentally a {\it classification} paradigm. Given
a description in terms of feature-value pairs of an input, a category
label is produced. This category should normally be taken from a
finite inventory of possibilities, known beforehand\footnote{This
restriction can be circumvented by having multiple classifiers predict
a different part of the output pattern, see \cite{ling} for this
approach in learning decision trees.}.  It is our hypothesis that {\it
all} useful linguistic tasks can be redefined this way. All linguistic
problems can be described as context-sensitive mappings.  These
mappings can be of two kinds: {\it identification} and {\it
segmentation} (identification of boundaries).

\begin{itemize}
\item {\bf Identification}. Given a set of possibilities (categories)
and a relevant context in terms of attribute values, determine the
correct possibility for this context. Instances of this include {\it
part of speech tagging}, {\it grapheme-to-phoneme conversion}, {\it
lexical selection in generation}, {\it morphological synthesis}, {\it
word sense disambiguation}, {\it term translation}, {\it stress
assignment}, etc.
\item {\bf Segmentation}. Given a target and a context, determine
whether and which boundary is associated with this target. Examples
include {\it syllabification}, {\it morphological analysis}, {\it
syntactic analysis} (in combination with tagging), etc.
\end{itemize}

An approach often necessary to arrive at the context information
needed is the {\it windowing} approach (as in \cite{sej:ros} for text
to speech), in which an imaginary window is moved one item at a time
over an input string where one item in the window (usually the middle
item or the last item) acts as a target item, and the rest as the
context.  An alternative possibility is to use {\em operators} as
categories, e.g., shift and different types of reduce as categories in
a shift-reduce parser (see \cite{simm} for such an approach outside
the context of Machine Learning).

\section{Examples}

The approach proposed in this paper is fairly recent, and experiments
have focused on phonological and morphological tasks rather than on
tasks like term disambiguation. However, we hope to have made clear
that the approach is applicable to all classification problems in NLP.
In this section we briefly describe some of the experiments and hope
the reader will refer to the cited literature for a more detailed
description.

\subsection{Syllable Boundary Prediction}

Here the task to be solved is to decide where syllable boundaries
should be placed given a word form in its spelling or pronunciation
representation (the target language was Dutch). In a knowledge-based
solution, we would implement well-known phonological principles like
the {\it maximal onset principle} and the {\it sonority hierarchy}, as
well as a {\it morphological parser} to decide on the position of
morphological boundaries, some of which overrule the phonological
principles. This parser requires at least lexical knowledge about
existing stems and affixes and the way they can be combined.

In the lazy learning approach (\cite{dae:bos:a};
\cite{dae:bos:b}), we used the windowing approach referred
to earlier to formulate the task as a classification problem (more
specifically, a segmentation problem). For each letter or phoneme, a
pattern was created with a target letter or phoneme, a left context
and a right context. The category was {\it yes} (if the target letter
or phoneme should be preceded by a syllable boundary) or {\it no} if
not. The lazy learning approach produced results which were more
accurate than both a connectionist approach (backpropagation learning
in a recurrent multi-layer perceptron) and a knowledge-based approach.
The information gain metric also ``discovered'' an interesting
asymmetry between predictive power of left and right context (right
context turned out to be more informative).

\subsection{Grapheme-to-Phoneme Conversion}

Grapheme-to-phoneme conversion is a central module in text-to-speech
systems. The task here is to produce a phonetic transcription given
the spelling of a word. Again, in the knowledge-based approach, the
lexical requirements for such a system are extensive. In a typical
knowledge-based system solving the problem, morphological analysis
(with lexicon), phonotactic knowledge, and syllable structure
determination modules are designed and implemented.

In a lazy learning approach (\cite{dae:bos:c}; \cite{bos:dae}), again
a windowing approach was used to formulate the task as a
classification problem (identification this time: given a set of
possible phonemes, determine which phoneme should be used to translate
a target spelling symbol taking into account its context).  Results
were highly similar to the syllable boundary prediction task: the lazy
learning approach resulted in systems which were more accurate than
both a connectionist approach and a linguistically motivated approach.
The results were replicated for English, French, and Dutch, using the
same lazy learning algorithm, which shows its reusability.

\subsection{Word Stress Assignment}

Another task we applied the lazy learning algorithm to, was stress
assignment in Dutch monomorphematic, polysyllabic words
(\cite{dae:al:a}, \cite{dae:al:b}). A word was coded by assigning one
feature to each part of the syllable structure of the last three
syllables (if present) of the word (see the description of the
diminutive formation task described earlier). There were three
categories: final stress, penultimate stress, and antepenultimate
stress (an identification problem).

Although this research was primarily intended to show that an
empiricist learning method with little a priori knowledge performed
better than a learning approach in the context of the ``Principles and
Parameters'' framework as applied to metrical phonology, the results
also showed that even in the presence of a large amount of noise (from
the point of view of the learning algorithm), the algorithm succeeded
in automatically extracting the major generalizations that govern
stress assignment in Dutch, with no linguistic a priori knowledge
except syllable structure.

\subsection{Part of Speech Tagging}

In this as yet unpublished research, a slightly more complex learning
procedure was applied to the problem of part of speech tagging (an
identification problem). First, a {\it lexicon} was derived from the
training set. The training set consists of a number of texts in which
each word is assigned the correct part of speech tag (its category).
To derive a lexicon, we find for each word how many times it was
associated with which categories. We can then make an inventory of
{\it ambiguous categories}, e.g., a word like {\it man} would belong
to the ambiguous category {\it noun-or-verb}. The next step consists
of retagging the training corpus with these ambiguous categories.
Advantages of this extra step are (i) that ambiguity is restricted to
what actually occurs in the training corpus (making as much use as
possible of sublanguage characteristics), and (ii) that we have a much
more refined measure of similarity in lazy learning: whereas
non-ambiguous categories can only be equal or not, ambiguous
categories can be {\em more or less} equal. For the actual tagging
problem, a moving window approach was again used, using patterns of
ambiguous categories (a target and a left and right context).  Results
are only preliminary here, but suggest a performance comparable to
hidden markov modeling approaches.

\section{Conclusion}

There are both theoretical and practical aspects to the work described
in this paper. First, as far as linguistic engineering is concerned, a
new approach to the reusability problem was proposed. Instead of
concentrating on linguistic engineering of theory-neutral,
poly-theoretic, multi-applicable lexical representations combined with
semi-automatic migration of lexical knowledge between different
formats, we propose an approach in which a single inductive learning
method is reused on different corpora representing useful linguistic
mappings, acquiring the necessary lexical information automatically
and implicitly.

Secondly, the theoretical claim underlying this proposal is that
language acquisition and use (and a fortiori lexical knowledge
acquisition and use) are behaviour-based processes rather than
knowledge-based processes. We sketched a {\it memory-based lexicon}
with the following properties:

\begin{itemize}
\item The lexicon is not a static data structure but a set of lexical
processes of identification and segmentation. These processes
implement lexical performance.
\item Each lexical process is represented by a set of exemplars
(solved cases) in memory, which act as models to new input.
\item New instances of a lexical process are solved through either
memory lookup or similarity-based reasoning.
\item There is no representational difference between regularities,
subregularities, and exceptions.
\item Rule-like behaviour is a side-effect of the operation of the
similarity matching process and the contents of memory.
\item The contents of memory (the lexical exemplars) can be approximated
as a set of rules for convenience.
\end{itemize}

In a broader context, the results described here argue for an
empiricist approach to language acquisition, and for exemplars rather
than rules in linguistic knowledge representation (see \cite{dae:al:b}
and Gillis et al. \cite{gill:al} for further discussion of these
issues).

There are also some limitations to the method. The most important of
these is the {\it sparse data problem}. In problems with a large
search space (e.g., thousands of features relevant to the task), a
large amount of training patterns is necessary in order to cover the
search space sufficiently. In general, this is not a problem in NLP,
where for most problems large corpora are available or can be
collected. Also, information gain or other feature weighting
techniques can be used to automatically reduce the dimensionality of
the problem, sometimes effectively solving the sparse data problem.

Another problem concerns {\it long-distance dependencies}, especially
in syntax. The methods described often make use of a moving window
approach in which only a local part of an input representation is
used.  Whenever important factors determining a category decision are
outside the scope of a pattern, the category assignment cannot be
learned. A possible solution for this problem is the cascading of
different lazy learning systems, one working on the output of the
other. For example, a learning system for part of speech tagging could
be combined with a learning system taking patterns of disambiguated
tags as input, and producing constituent types as output. Taking
patterns of constituent types as input, a third learning system should
have no problem assigning ``long-distance'' dependencies: given the
right representation, all dependencies are local.

\end{document}